\documentclass[a4paper]{article}

\usepackage{INTERSPEECH2020}
\usepackage{hyperref}

\usepackage{xcolor}
\usepackage[capitalize]{cleveref}

\newcommand{\RR}{\mathbb{R}}

\newcommand{\GN}{\mathcal{N}}
\newcommand{\eone}{\textsc{(E1)}}
\newcommand{\etwo}{\textsc{(E2)}}
\newcommand{\etwoa}{\textsc{(E2A)}}
\newcommand{\etwob}{\textsc{(E2B)}}

\newcommand{\norm}[1]{\ensuremath{\lVert #1 \rVert}}

\title{Analyzing Robustness of End-to-End Neural Models for Automatic Speech Recognition}
\name{Goutham Rajendran $^*$, Wei Zou \sthanks{$^*$ Equal Contribution}}
\address{
  University of Chicago}
\email{goutham@uchicago.edu, weizou@uchicago.edu}

\begin{document}
\maketitle

\begin{abstract}
    We investigate robustness properties of pre-trained neural models for automatic speech recognition. Real life data in machine learning is usually very noisy and almost never clean, which can be attributed to various factors depending on the domain, e.g. outliers, random noise and adversarial noise. Therefore, the models we develop for various tasks should be robust to such kinds of noisy data, which led to the thriving field of robust machine learning. We consider this important issue in the setting of automatic speech recognition. With the increasing popularity of pre-trained models, it's an important question to analyze and understand the robustness of such models to noise.
    In this work, we perform a robustness analysis of the pre-trained neural models wav2vec2, HuBERT and DistilHuBERT on the LibriSpeech and TIMIT datasets. We use different kinds of noising mechanisms and measure the model performances as quantified by the inference time and the standard Word Error Rate metric. We also do an in-depth layer-wise analysis of the wav2vec2 model when injecting noise in between layers, enabling us to predict at a high level what each layer learns. Finally for this model, we visualize the propagation of errors across the layers and compare how it behaves on clean versus noisy data. Our experiments conform the predictions of Pasad et al. [2021] and also raise interesting directions for future work.
\end{abstract}

\section{Introduction}
    Speech recognition has undergone a revolution with the success of pre-trained models. Pre-trained models such as wav2vec2 \cite{baevski2020wav2vec} amd HuBERT \cite{hsu2021hubert} are growing in popularity and are being used widely for a variety of speech-related tasks. With this unprecedented growth, natural issues should be considered. One such issue is the measure of robustness of the model. Robustness of a model can informally be defined to be the amount of data noise that the model can handle without dimishing too much in accuracy.
    The study of robustness has had important applications in many fields of machine learning especially in computer vision, since they have safety-critical relevance to downstream tasks like autonomous driving.
    In speech recognition, data is almost never noise-free, so noise is baked in. Indeed, any realistic speech signal has background noise. Moreover, noise can occur in various other forms stemming from hardware and software issues, such as white noise, corrupted frames, etc.

    In this work, we analyze the performance of popular pre-trained neural speech models with an eye towards such issues.
    We perform two classes of experiments. In the first class, we directly inject noise to the raw waveform input and study model performance, as quantified by inference time and the Word Error Rate. In the second class of experiments, we inject noise in between layers of the neural model during inference and study the model behavior. This experiment offers deeper insight into the layer-wise behavior of the neural models. Intuitively, layers that learn higher order semantic information about the speech should be more robust to noise. For instance, Pasad et al. \cite{pasad2021layer} predict that some layers of the wav2vec 2.0 model likely encode phonetic information, while some others likely encode higher level information such as word content or context. This suggests that the layers that learn higher level contextual information should be more robust to noise. Our final experiment is to visualize and compare the activations of the layers on uncorrupted and corrupted data, to see if and when the noise is eliminated in the neural network. This enables us to better understand the representations that the model learns.

    We now describe these two classes of experiments in more detail. In the first class $\eone$, we study the performance of the models on noisy data, where we quantify performance by inference time and the standard Word Error Rate metric. We experiment with various types of simulated noise, including white noise, speed perturbation and dropped frame chunks. In particular, for all these kinds of noise, we compare fine-tuned wav2vec2 \cite{baevski2020wav2vec} and HuBERT \cite{hsu2021hubert} models on the LibriSpeech dataset \cite{panayotov2015librispeech}; and also compare fine-tuned wav2vec2 and DistilHuBERT \cite{chang2022distilhubert} models on the TIMIT dataset \cite{garofolo1993darpa}.

    In the second class of experiments $\etwo$, we focus on wav2vec2 and do a  more detailed layer-wise analysis. In our first experiment of this class $\etwoa$, during inference on data, we intervene in a specific layer and inject (white) noise. We then let the inference proceed as usual. We repeat this for other layers and study how the model performance degrades with our intervention. We repeat this experiment for both additive and multiplicative noise. In our second experiment $\etwob$, we do model inference with the original and noisy data and compare how the activations differ in each layer.

    Our findings are as follows
    \begin{enumerate}
        \item From $\eone$, we conclude that HuBERT is (around $25\%$) slower than wav2vec2 but on simulated noise, it's more robust than wav2vec2. Similarly, DistilHuBERT is more robust than wav2vec2.
        \item In the additive noise version of $\etwoa$ we observe that layers 6-8 seem very noise sensitive while other layers seem relatively noise-robust. This suggests that layers 6-8 learn higher level information such as semantics, context or meaning. This matches the observations of \cite{pasad2021layer}.
        \item Almost all layers of wav2vec2 are surprisingly robust on multiplicative noise injection of $\etwob$, except for layer $11$ which behaves in a unusual manner.
        \item The experiments $\etwob$ suggests that the wav2vec2 model ``eats'' up the noise as we go up the layers. This is true for all layers except layer 11, where intriguing things seem to happen.
    \end{enumerate}
\section{Related work}

Robust speech recognition has always been an important research avenue in speech technologies, e.g. the works \cite{sadhu2021wav2vec, hsu2021robust, wang2022wav2vec, ravanelli2020multi, wang2020complex, subramanian2019speech}, also see the book \cite{li2015robust}. Perhaps, the work most closely related to ours is \cite{lin2022analyzing} where they analyze model robustness on mismatched domains. Many of these prior modeling works attempt to build models that are robust to noise. Towards this goal, they use various techniques such as by having a dedicated denoiser module or a speech enhancement module or make the model robust enough so that it learns to discard such noise from the data while it's processing the data.
In the latter technique, a standard approach is via data augmentation. In this approach, we augment the dataset with noisy data and then fine-tune it. It's known that this reduces generalization error,  e.g. see \cite{shorten2019survey}. For prior works studying this in the context of speech, see e.g. \cite{braun2017curriculum, hsu2021robust}. Finally, some works have also explored adversarial noise \cite{alzantot2018did, neekhara2019universal, olivier2022recent}.

\section{Experiments and Findings}

To aid our experiments, we use SpeechBrain \cite{speechbrain} and open-source versions of various datasets and pre-trained models from HuggingFace. In particular, we use the fine-tuned wav2Vec2-Base-960h model and the fine-tuned Hubert-Large-ls960 model.
For LibriSpeech, we use a sub-sampled version of the test-clean split and for TIMIT, we use a sub-sampled version of the test split. Experiments were run on an NVIDIA 1080i GPU with 64 GB memory. Our code is available at \url{https://github.com/weizou52/Robustness_Analysis_ASR}.

\subsection{Experiment \eone - Noisy waveform input}

In this section, we perturb the raw input with various kinds of noise and study the behavior of the model. More precisely, let $x \in \RR^n$ be the input raw waveform. We then add noise to it to obtain $x' = f(x) \in \RR^{n}$. We will explore various kinds of noise.

For the LibriSpeech dataset, we compare the word error rates of wav2vec2 and HuBERT. We conclude that HuBERT is more robust than wav2vec2 for this kind of simulated noise. Although it's worth remarking that inference time for HuBERT is also slower than wav2vec2. All our plots show the average inference time per datapoint.

\subsubsection{White Noise}

\begin{figure}[!ht]
    \centering
    \includegraphics[scale=.25, trim={0 0 0 0},clip]{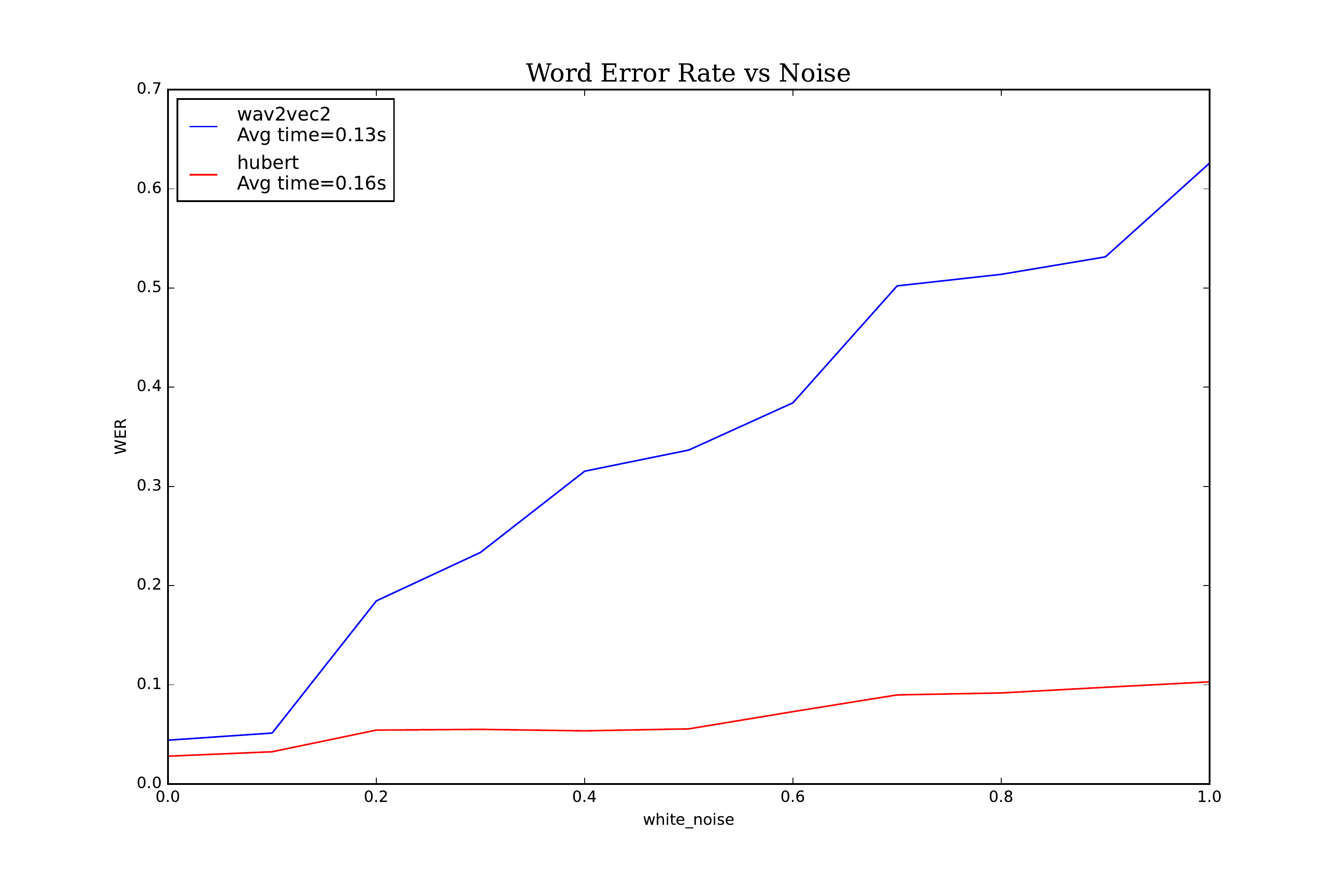}
    \caption{Word error rate as a function of $\rho$ (White noise)}
    \label{whitenoise}
\end{figure}

White noise is when we take the input $x \in \RR^n$, sample a random $g \sim \GN(0, I_n)$ and independently for each coordinate $i \le n$, with probability $\rho$, we set $x_i = x + g_i$, otherwise we don't change $x_i$. Here, $\rho$ is called the mixing probability.
This is the simplest form of noise we could add and helps set a baseline for further experiments. In \cref{whitenoise}, we plot our results.

\subsubsection{Speed Perturb}

\begin{figure}[!ht]
    \centering
    \includegraphics[scale=.25, trim={0 0 0 0},clip]{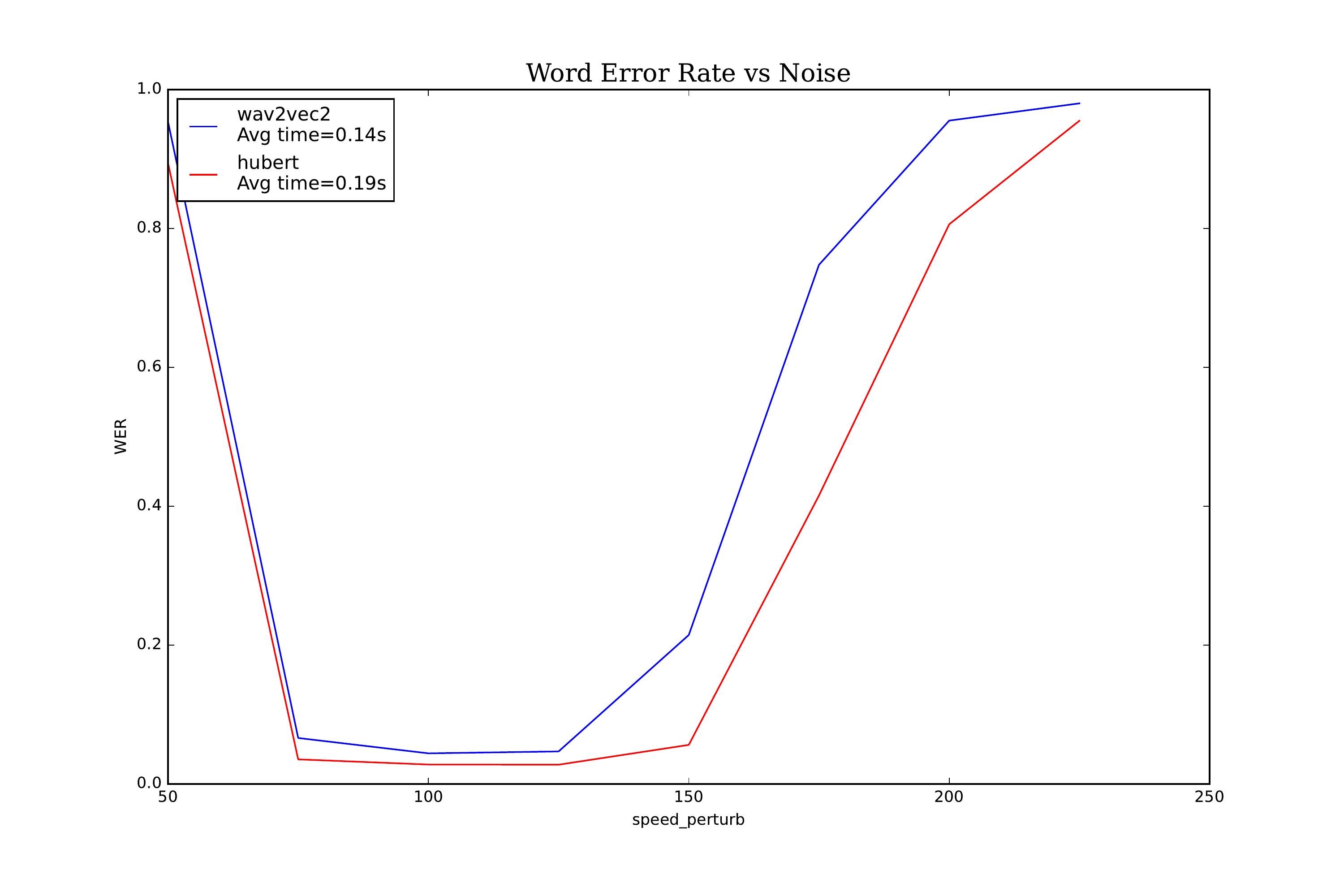}
    \caption{Word error rate as a function of $\rho$ (Speed Perturbation)}
    \label{speedperturb}
\end{figure}

Speed perturbation speeds up or slows down the speech. For a given speech signal $x$ and speed $100/\rho$, $f(x)$ is computed by resampling the audio signal without changing the sampling rate, using the technique in \cite{ko2015audio}. See \cref{speedperturb} for the results. The plot conforms with our intuition that speech that is sped up or slowed down is harder to predict, giving rise to the convex-looking plot.

\subsubsection{Chunk drop}

\begin{figure}[!ht]
    \centering
    \includegraphics[scale=.25, trim={0 0 0 0},clip]{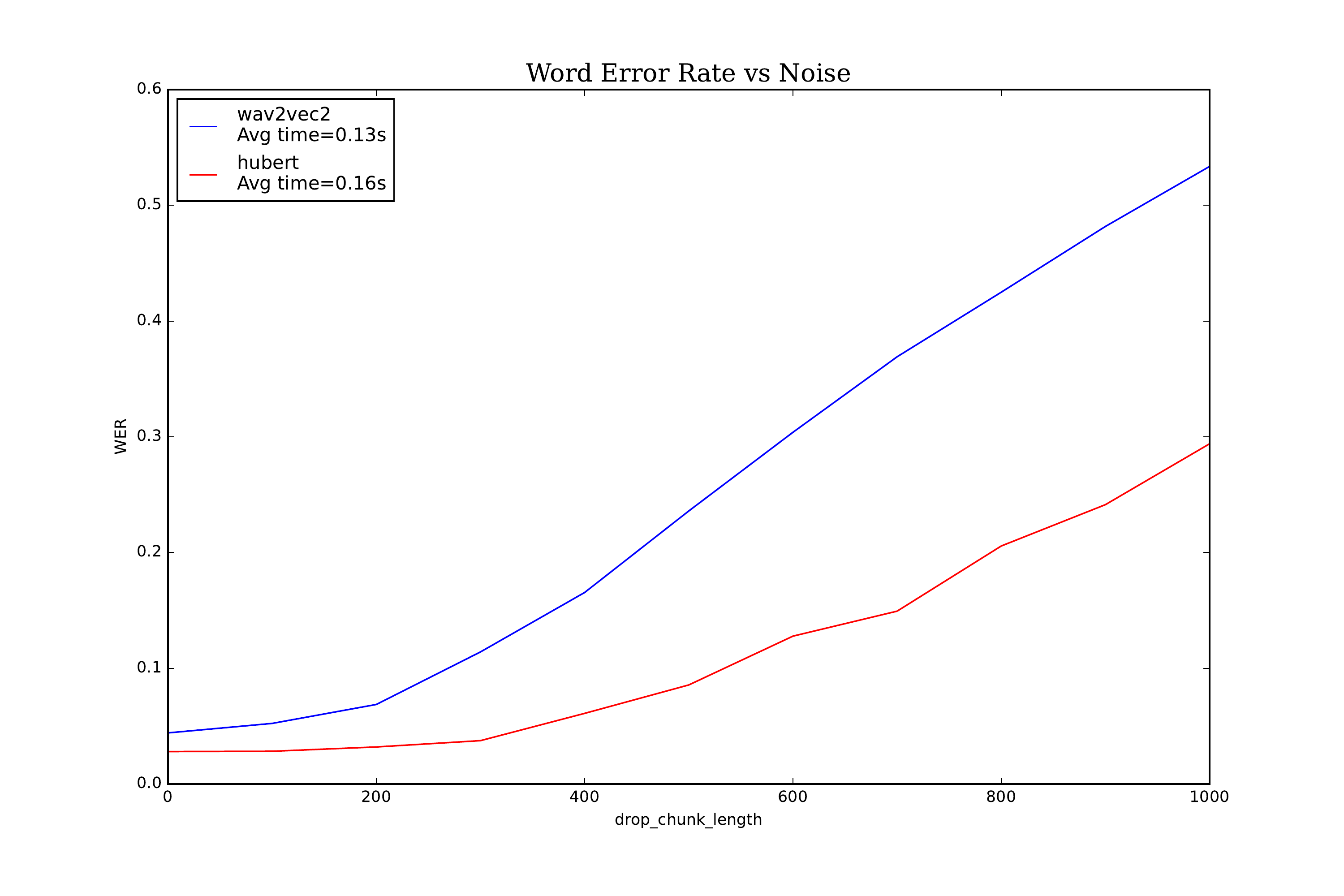}
    \caption{Word error rate as a function of $l$ (Dropping chunks)}
    \label{dropchunklength}
\end{figure}

\begin{figure}[!ht]
    \centering
    \includegraphics[scale=.25, trim={0 0 0 0},clip]{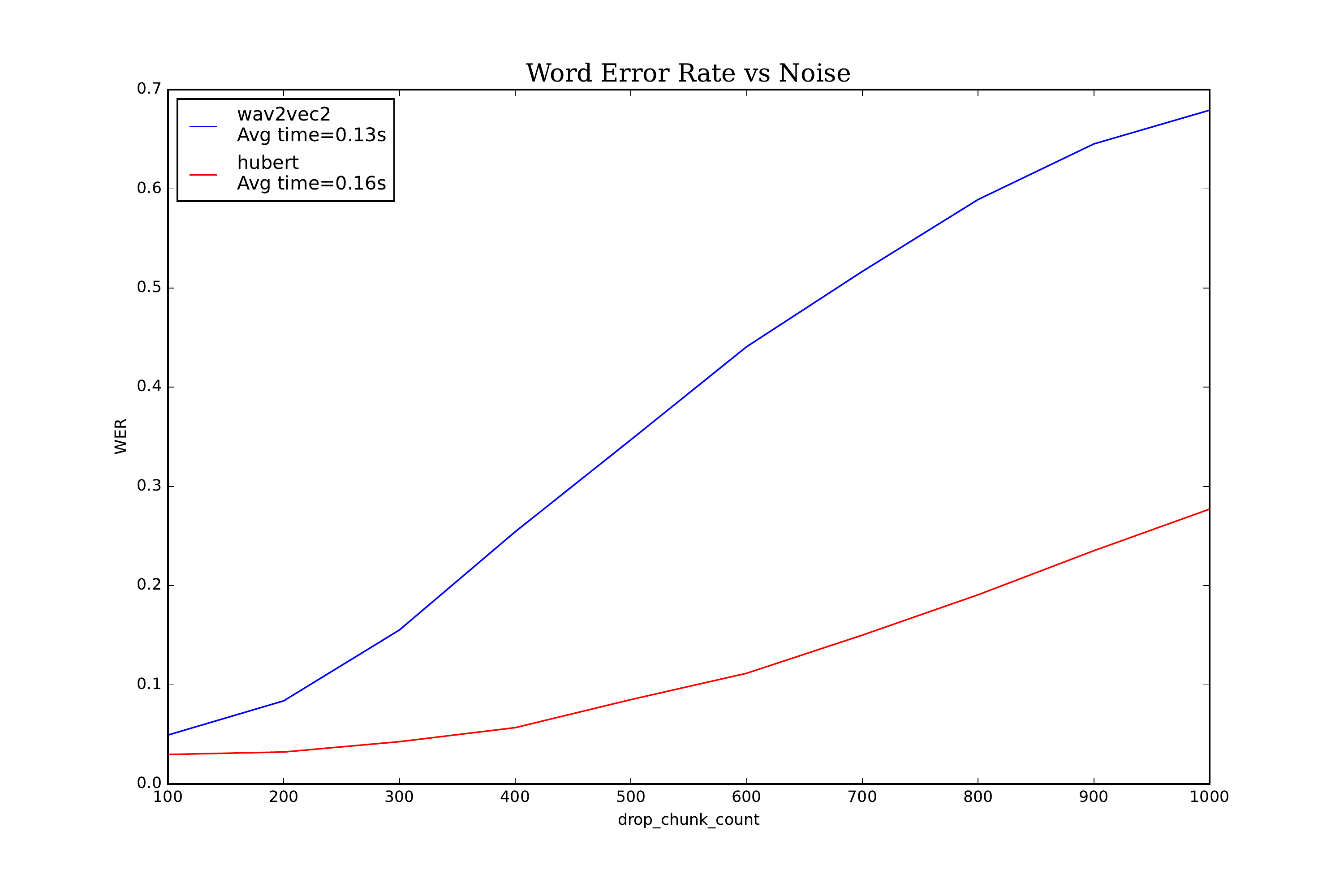}
    \caption{Word error rate as a function of $k$ (Dropping chunks)}
    \label{dropchunknumber}
\end{figure}

In our next experiment, we drop portions of our input signal $x$. We experiment both with number of chunks dropped $k$ and the length of each chunk that's dropped $l$. For the results of WER vs chunk length $l$ when number of dropped chunks $k$ is fixed to be $100$, see \cref{dropchunklength}. For the results of WER vs number of dropped chunks $k$ when chunk length $l$ is fixed to be $100$, see \cref{dropchunknumber}.

\subsection{Experiments on TIMIT}

\begin{figure}[!ht]
    \centering
    \includegraphics[scale=.25, trim={0 0 0 0},clip]{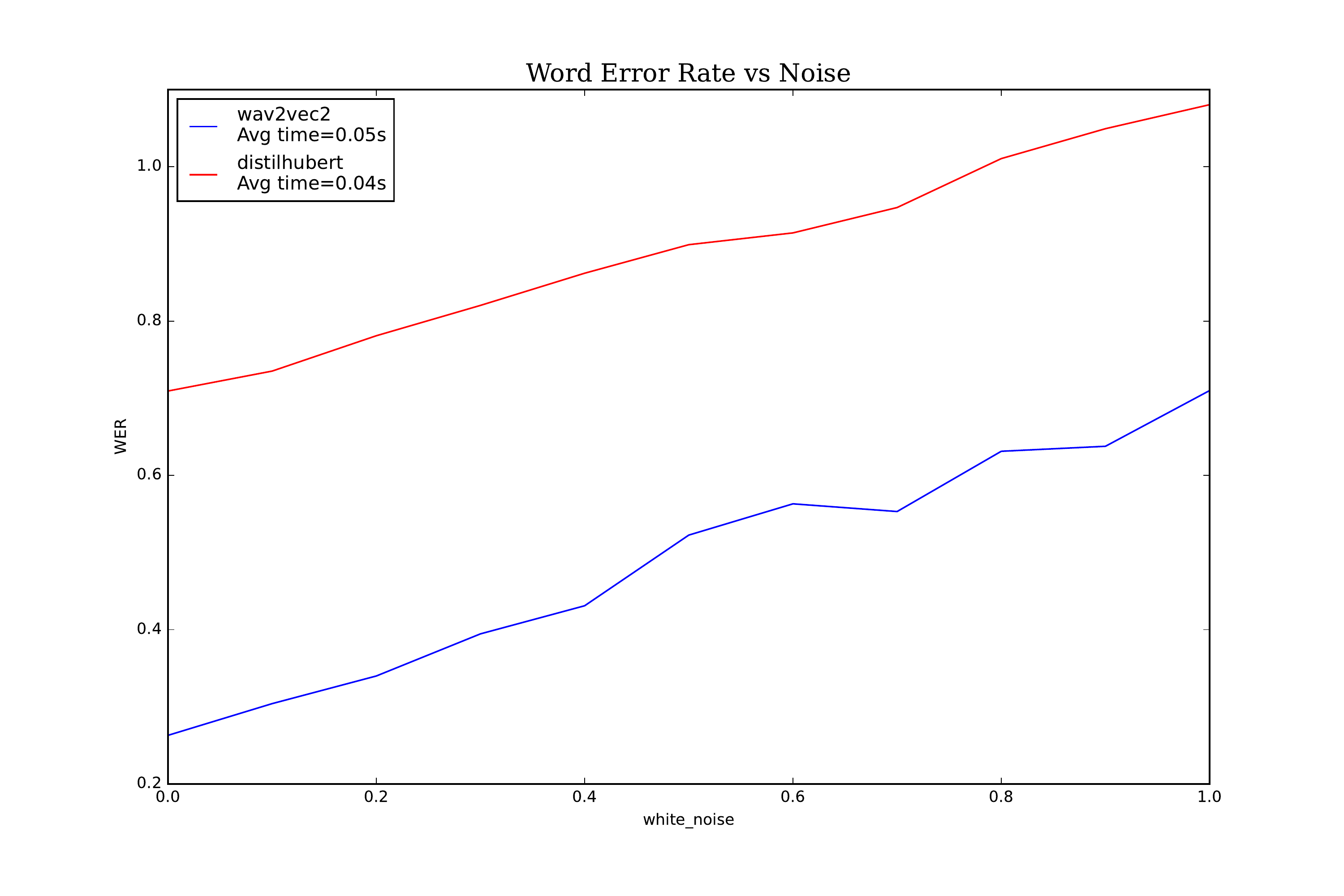}
    \caption{Word error rate as a function of $\rho$ (White noise)}
    \label{timit_whitenoise}
\end{figure}

\begin{figure}[!ht]
    \centering
    \includegraphics[scale=.25, trim={0 0 0 0},clip]{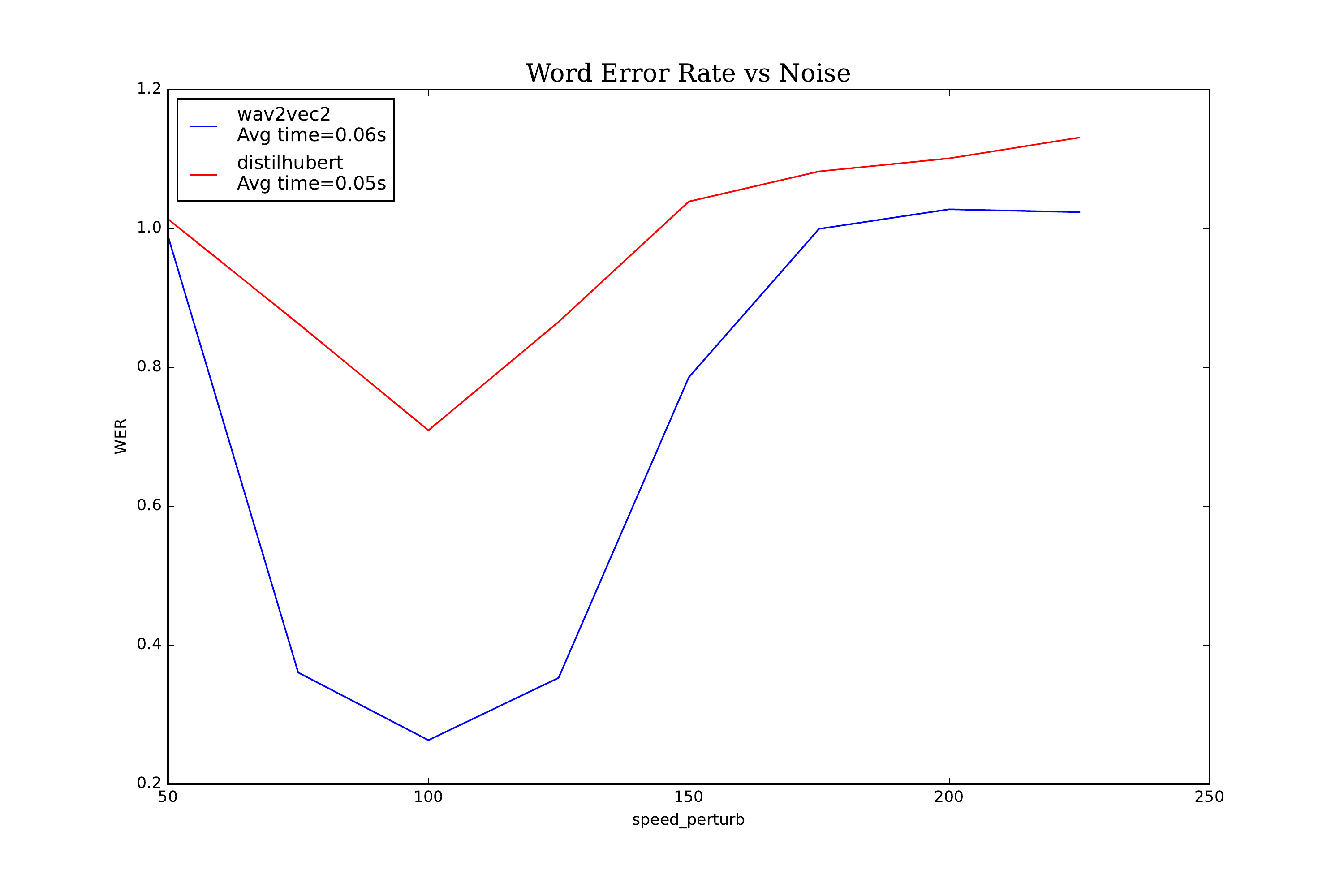}
    \caption{Word error rate as a function of $\rho$ (Speed Perturbation)}
    \label{timit_speedperturb}
\end{figure}

\begin{figure}[!ht]
    \centering
    \includegraphics[scale=.25, trim={0 0 0 0},clip]{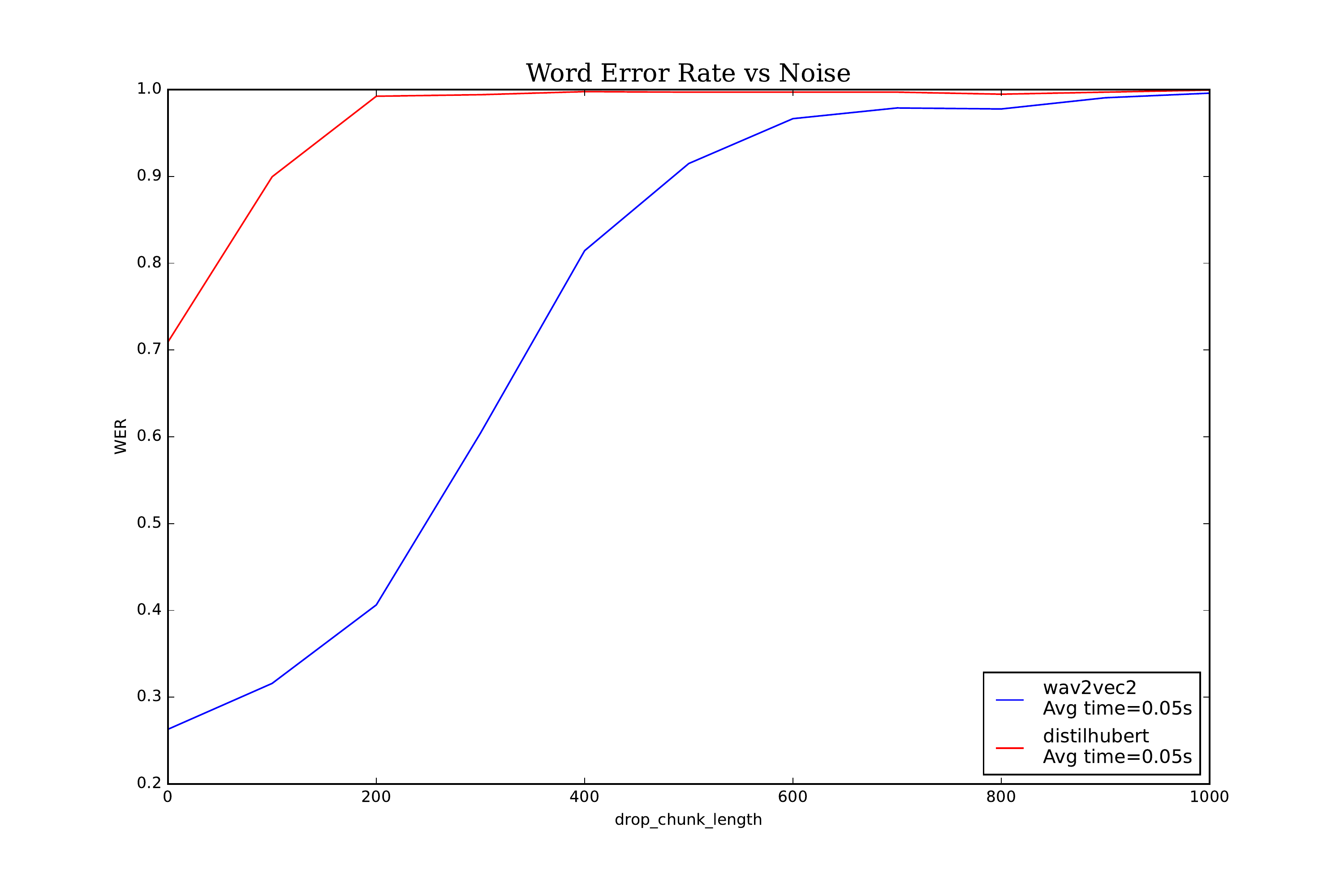}
    \caption{Word error rate as a function of $l$ (Dropping chunks)}
    \label{timit_dropchunklength}
\end{figure}

\begin{figure}[!ht]
    \centering
    \includegraphics[scale=.25, trim={0 0 0 0},clip]{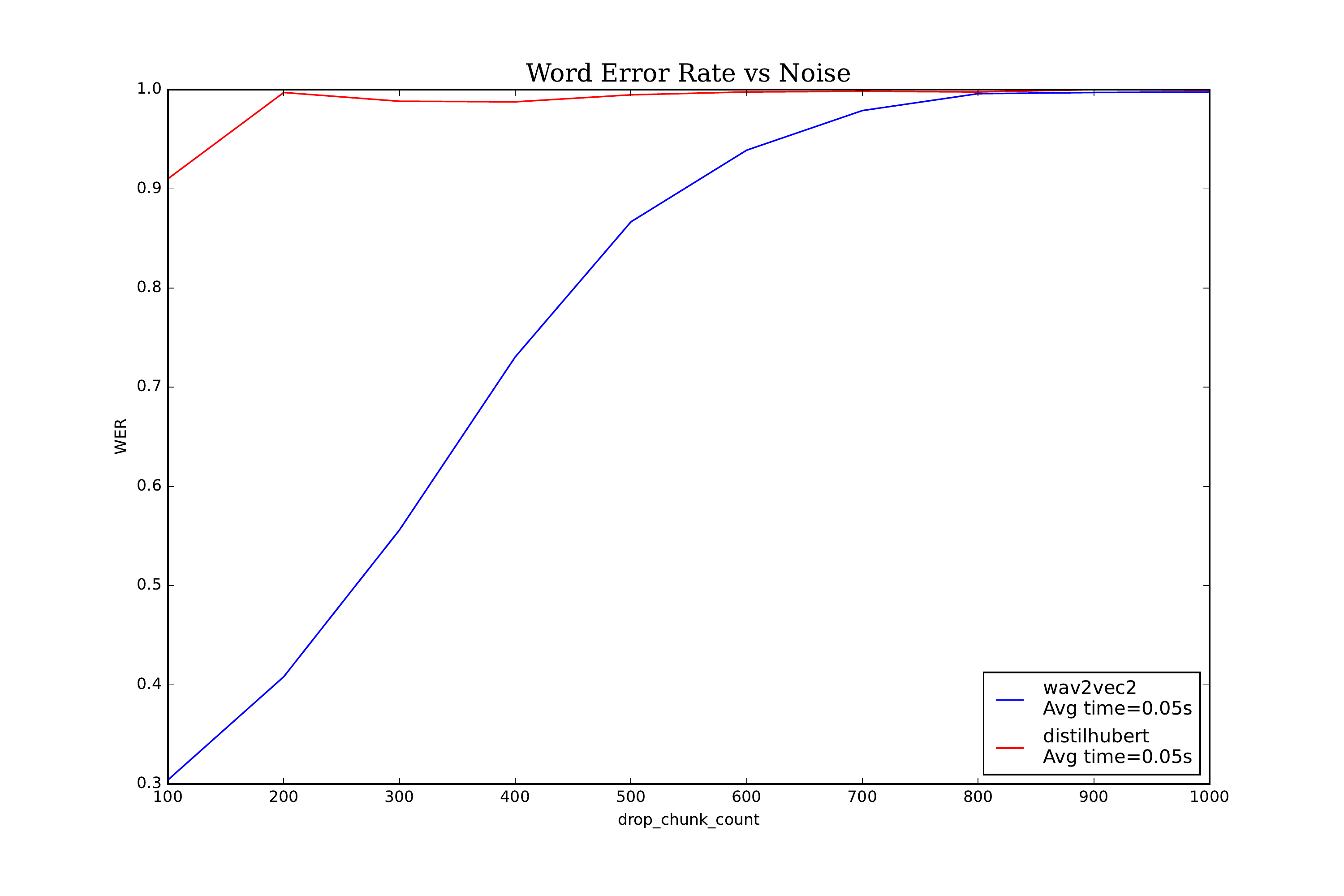}
    \caption{Word error rate as a function of $k$ (Dropping chunks)}
    \label{timit_dropchunknumber}
\end{figure}

We repeat these experiments on the TIMIT dataset where we compare wav2vec2 and DistilHuBERT. The plots obtained are in \cref{timit_whitenoise}, \cref{timit_speedperturb}, \cref{timit_dropchunklength} and \cref{timit_dropchunknumber}. In particular, note that since TIMIT is traditionally a dataset meant for phone recognition and moreover the diversity in the dictionary isn't very high, the word error rate metrics aren't remarkable. Nevertheless, the trend of DistilHuBERT outperforming wav2vec2 in robustness can still be seen.

\subsection{Experiment \etwo - Layerwise analysis}

In this section, we analyze the layers of wav2vec2 in different ways to gain more insight into the robustness properties of the model.

\subsubsection{Experiment \etwoa - Injecting noise into the layers}

\begin{figure}[!ht]
    \centering
    \includegraphics[scale=.25, trim={0 0 0 0},clip]{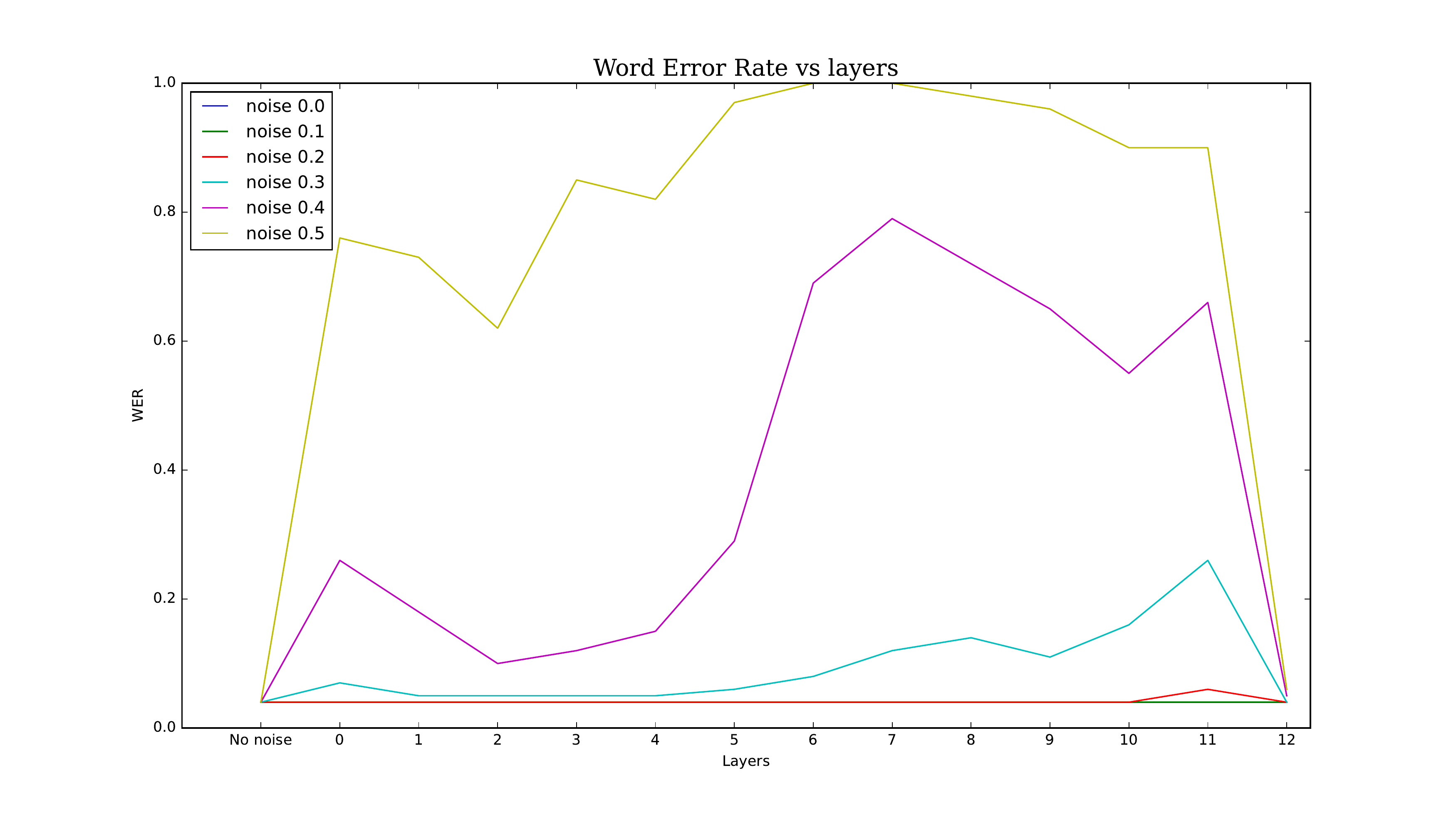}
    \caption{WER against layer (where additive noise is injected)}
    \label{werlayers}
\end{figure}

\begin{figure}[!ht]
    \centering
    \includegraphics[scale=.25, trim={0 0 0 0},clip]{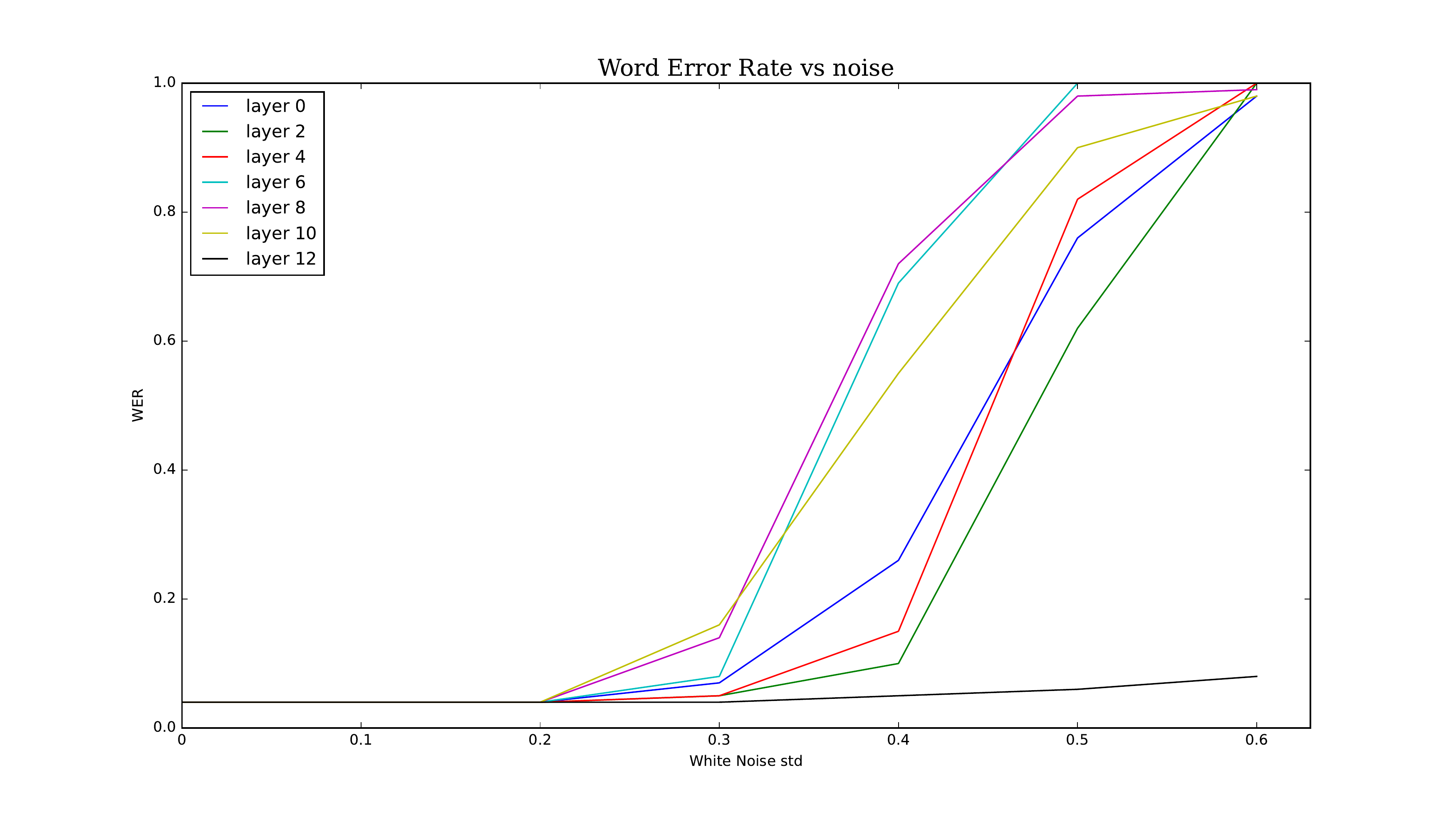}
    \caption{WER against $\rho$ (additive noise)}
    \label{wernoise}
\end{figure}

While performing inference on the test dataset, we inject noise in between various layers and let inference proceed as usual. Specifically, let the output of layer $i$ be $out_i \in \mathbb{R}^{d_i}$ where $d_i$ is the total number of activations of the $i$-th layer. For a fixed value of $i$, we modify this output \begin{equation}\label{add_noise}
    out_i' = out_i + \rho \cdot g
\end{equation}
where $g \sim \GN(0, I_{d_i})$ and $\rho$ is the standard deviation. In each run, we only noise one layer and the other layers are unaffected. By layer $0$, we mean the CNN feature extractor, just as in \cite{pasad2021layer}. In \cref{werlayers}, we show the changes in WER with respect to layers (where the noise $\rho \cdot g$ is injected). The same data is shown in \cref{wernoise} but as a function of WER with respect to $\rho$ for every other layer.

Because layers 6-8 seem fairly sensitive to noise compared to other layers which are noise robust, we conclude that layers 6-8 learn higher level information about the speech signal, such as context, meaning and semantics. Whereas other layers learn local information such as phone information or task specific information. The intuition is that local information or lower level information is robust to noise because surrounding contexts can help denoise but on the other hand, higher level information is sensitive to noise. This conforms with the findings of \cite{pasad2021layer}.


\begin{figure}[!ht]
    \centering
    \includegraphics[scale=.25, trim={0 0 0 0},clip]{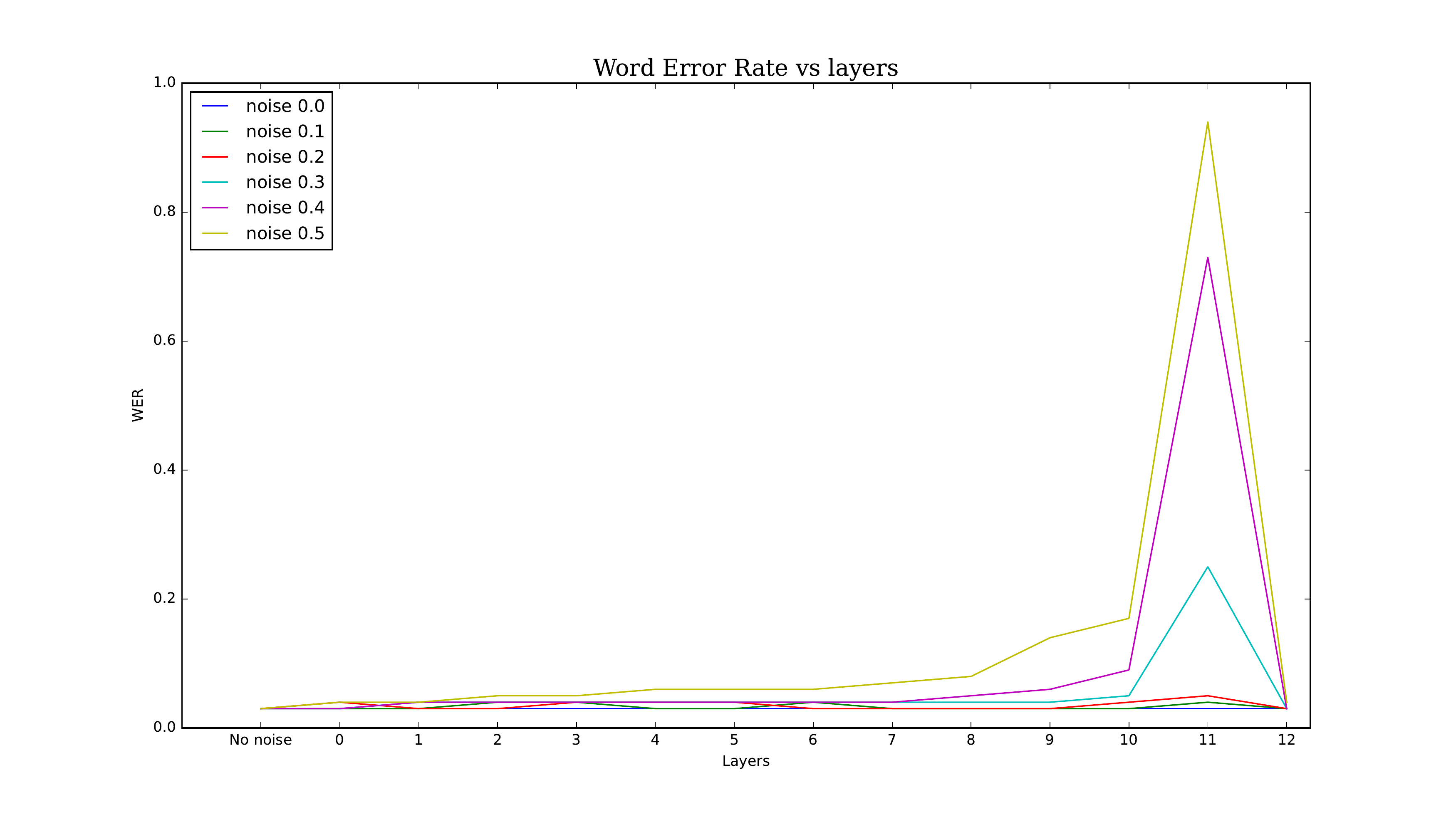}
    \caption{WER against layer (where multiplicative noise is injected)}
    \label{multwerlayers}
\end{figure}

\begin{figure}[!ht]
    \centering
    \includegraphics[scale=.25, trim={0 0 0 0},clip]{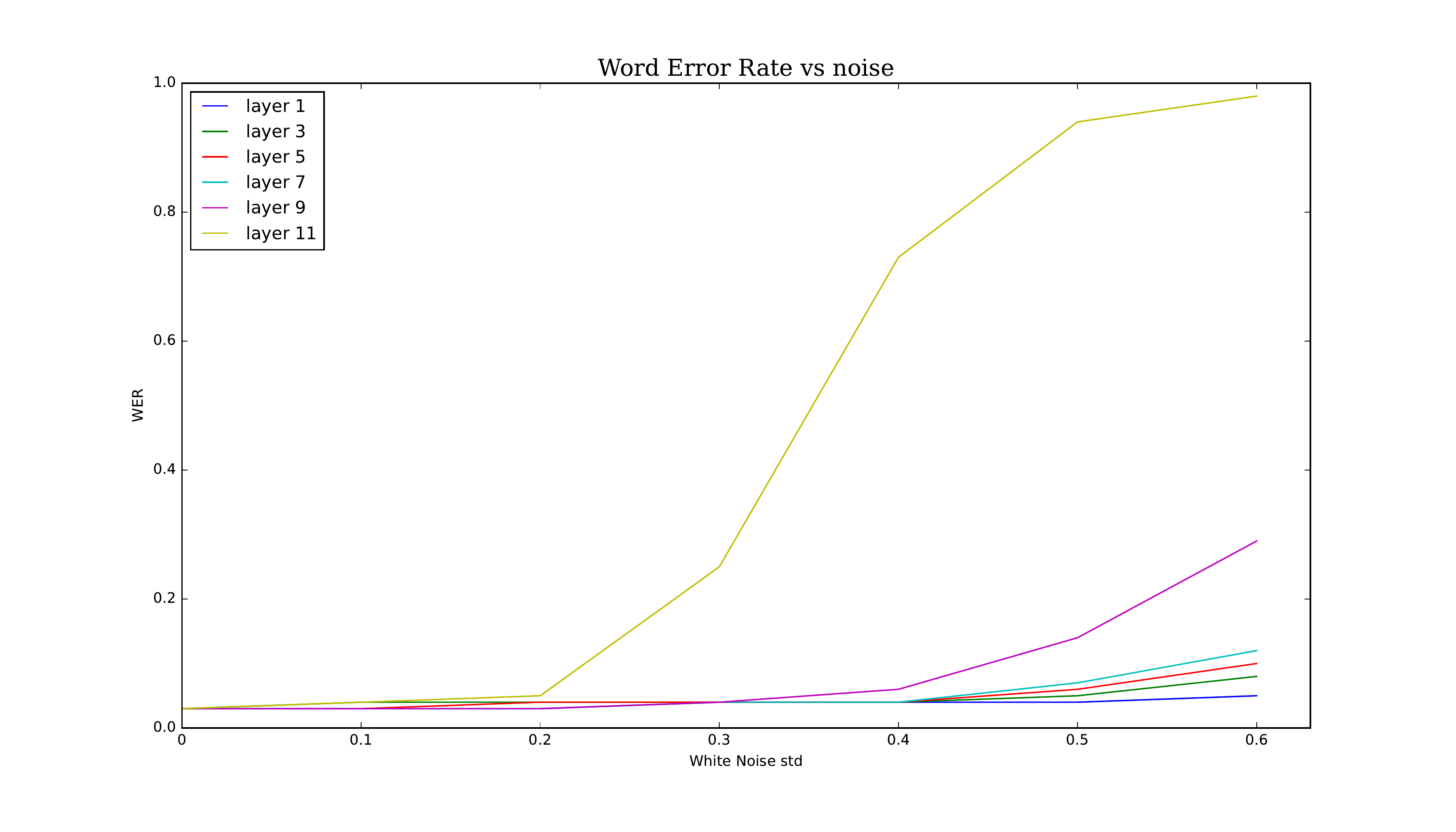}
    \caption{WER against $\rho$ (multiplicative noise)}
    \label{multwernoise}
\end{figure}

The kind of noise in \cref{add_noise} may not be the right kind of noising since different layers may have different scales. Therefore, we also consider the following kind of noising which takes scaling into account. Let $out_{i, j}$ be the $j$th coordinate of $out_i$ for $j \le d'$. Then, we noise as follows
\begin{equation}
    out_{i, j}' = out_{i, j}(1 + \rho \cdot g_j)
\end{equation}
where $g_j$ are iid sampled from $\GN(0, 1)$. For this kind of multiplicative noise, the corresponding outputs are shown in \cref{multwerlayers} and \cref{multwernoise}. In particular, note that layer $11$ is highly sensitive to noise. Similar unusual behavior of layer $11$ was observed in \cite{pasad2021layer}.


\subsubsection{Experiment \etwob - Evolution of the activations on noisy data}

\begin{figure}[!ht]
    \centering
    \includegraphics[scale=.25, trim={0 0 0 0},clip]{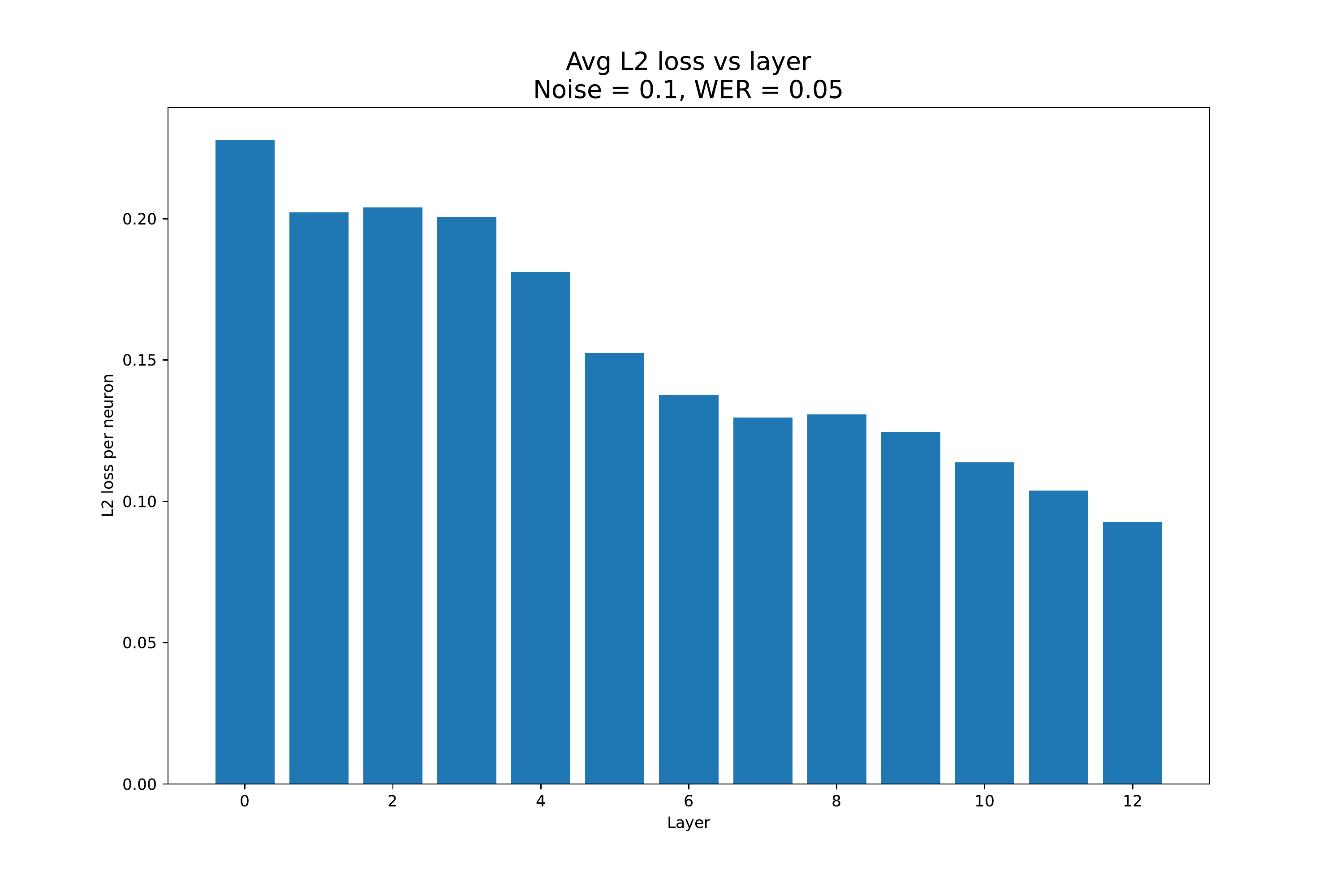}
    \caption{Average L2 loss across layers with $\rho = 0.1$}
    \label{l2losslayers}
\end{figure}

In our next experiment, we compare how the model inference propogates across layers when the inputs are $x^{(1)}$ and $x^{(2)} = x^{(1)} + \rho \cdot g$ where $g \sim \GN(0, I_n)$. In particular, for layer $i$, we compute the normalized L2 loss

\begin{equation}
    dist_i = \frac{1}{\sqrt{d_i}}\norm{out_i^{(1)} - out_i^{(2)}}_2
\end{equation}
where $out_i^{(1)}$ and $out_i^{(2)}$ are the activations of layer $i$ on inputs $x^{(1)}$ and $x^{(2)}$ respectively, and $d_i$ is the number of neurons in the $i$th layer. To compute the L2 loss per neuron, we divide by the scaling factor $\sqrt{d_i}$. Finally, we take the average over the samples from the test dataset. In \cref{l2losslayers}, we plot the loss as we go higher up in the layers. In this plot, we fix $\rho = 0.1$. As we can see, the layers are essentially nullifying the loss incurred as we go higher up in the model, suggesting that when the model performs inference, it's simultaneously denoising.

\begin{figure}[!ht]
    \centering
    \includegraphics[scale=.25, trim={0 0 0 0},clip]{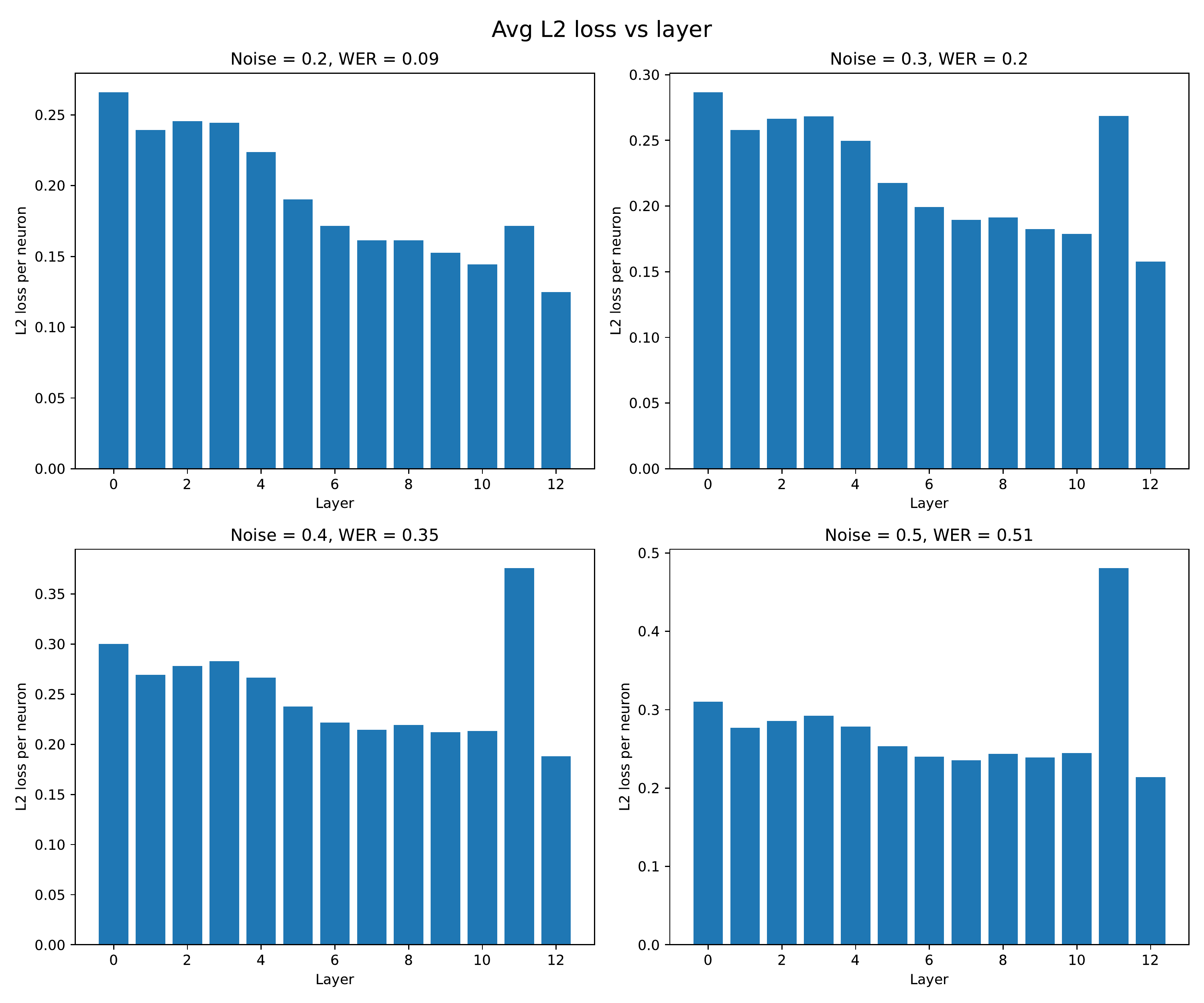}
    \caption{Average L2 loss across layers for various $\rho$}
    \label{layer11}
\end{figure}

In \cref{layer11}, we repeat the experiment for different values of $\rho$. We notice a similar trend except for the unusual layer $11$. Strange behavior for layer $11$ was also observed in \cite{pasad2021layer} and it's possible that they are related, whose investigation we leave for future work.

\section{Potential future directions}

\begin{enumerate}
    \item Background noise: Background noise is an interesting direction for further study, which perhaps maybe a more realistic kind of noise. To generate realistic background noise, some works, e.g. \cite{wang2022wav2vec} mixed other speech datasets with the main one, using various techniques. So it would be interesting to repeat our experiments in this setting.
    \item Adversarial noise: Compared to a field like Computer Vision, there is very limited research exploring adversarial noise in speech (e.g. \cite{alzantot2018did, neekhara2019universal}) but \textit{all these works are in the pre-transformer era}.
    But it seems like in recent months, the Speech community is slowly starting to focus their attention on adversarial noise, e.g. \cite{olivier2022recent}. So this is a deeply fascinating area for further research.
\end{enumerate}

\section{Acknowledgements}

We thank Ju-Chiech Chou and Karen Livescu for useful comments and suggestions. We thank Toyota Technological Institute at Chicago for their compute clusters.

\bibliographystyle{IEEEtran}

\bibliography{mybib}

\end{document}